%% file: main.tex
\documentclass[runningheads]{llncs}
\usepackage[T1]{fontenc}
\usepackage{graphicx}
\usepackage{booktabs}
\usepackage[misc]{ifsym}
\newcommand{\corr}{(\Letter)}
\usepackage{mwe}

\input{preamble}

\begin{document}

\toctitle{Subspace Optimization for Backpropagation-Free Continual Test-Time Adaptation}
\tocauthor{Damian~Sójka,Sebastian~Cygert,Marc~Masana}

\title{Subspace Optimization for Backpropagation-Free\\Continual Test-Time Adaptation}

\titlerunning{Subspace Optimization for Backpropagation-Free Continual TTA}

\author{Damian Sójka\inst{1,2} \corr \and
Sebastian Cygert\inst{3,4,5} \and
Marc Masana\inst{6}}

\authorrunning{D.~Sójka et al.}

\institute{\leavevmode
  \inst{1} Poznan University of Technology \quad 
  \inst{2} Akces NCBR \\ 
  \inst{3} NASK - National Research Institute \quad 
  \inst{4} Gdańsk University of Technology \\ 
  \inst{5} IDEAS Research Institute \quad
  \inst{6} Graz University of Technology \\
  \email{damian.sojka@doctorate.put.poznan.pl}
}

\maketitle              

\sloppy

\begin{abstract}
We introduce \methodname{}, a backpropagation-free continual test-time adaptation system that directly optimizes the affine parameters of normalization layers. Existing derivative-free approaches struggle to balance runtime efficiency with learning capacity, as they either restrict updates to input prompts or require continuous, resource-intensive adaptation regardless of domain stability. To address these limitations, \methodname{} leverages the Covariance Matrix Adaptation Evolution Strategy with the Fastfood projection to optimize high-dimensional affine parameters within a low-dimensional subspace, leading to superior adaptive performance. Furthermore, we enhance the runtime efficiency by incorporating an adaptation stopping criterion and a domain-specialized vector bank to eliminate redundant computation. Our framework\footnote{Code available at~\url{https://github.com/dmn-sjk/PACE.git}. Accepted for publication at ECML PKDD 2026. The final authenticated Version of Record will be published by Springer in the LNCS series.} achieves state-of-the-art accuracy across multiple benchmarks under continual distribution shifts, reducing runtime by over 50\% compared to existing backpropagation-free methods.
\end{abstract}


\section{Introduction}
\label{sec:intro}
\begin{figure}[t]
    \centering
    \includegraphics[width=\linewidth]{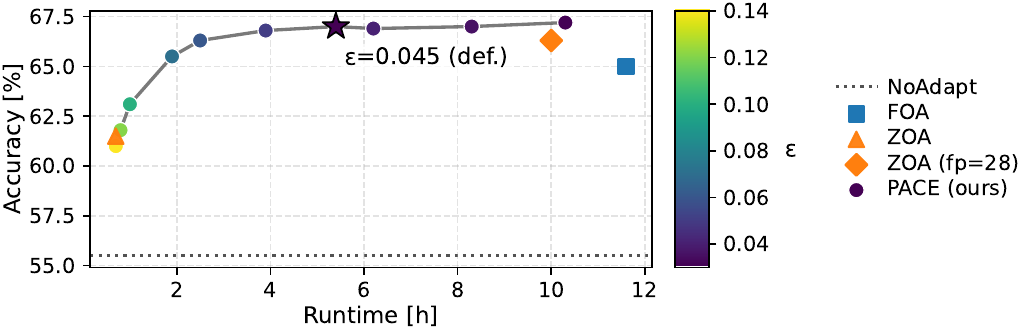}
    \caption{
    \textbf{Accuracy versus runtime trade-off} on the \mbox{ImageNet-C} benchmark using a \mbox{ViT-B} model, across various adaptation stopping thresholds $\epsilon$ (star marks the default setting $\epsilon\!=\!0.045$). 
    The horizontal dotted line represents the NoAdapt baseline accuracy. Existing BP-free methods typically face a trade-off: they either achieve high accuracy at the expense of computational efficiency or reduce runtime by sacrificing precision. Our approach not only outperforms current baselines but also introduces a tunable mechanism to balance the accuracy-runtime trade-off, preventing the inefficient use of resources for diminishing returns.
    }
    \label{fig:performance_whole_inc}
\end{figure}

Test-time adaptation~(TTA)~\cite{wang2021tent} has emerged as a practical approach to adapt a deployed neural networks on-the-fly, increasing its robustness to shifting data distributions. While backpropagation~(BP)-based methods~\cite{niu2023towards,eata,wang2022continual,wang2021tent} achieve strong performance via self-supervised learning, their high memory requirements and incompatibility with non-differentiable quantized models limit their deployment on resource-constrained edge devices~\cite{niu2024test,deng2025test,jia2024tinytta}. Conversely, although existing BP-free methods reduce memory overhead and support quantization~\cite{boudiaf2022parameter,iwasawa2021test,deng2025test,niu2024test}, the inherent challenges of derivative-free adaptation hinder their ability to balance runtime efficiency with learning capacity~(Fig.~\ref{fig:performance_whole_inc}).

Existing BP-free methods suffer from two primary limitations:
\ 1) restricted learning capacity: current state-of-the-art does not update the model parameters~\cite{boudiaf2022parameter, iwasawa2021test, schneider2020improving, khurana2021sita, lim2023ttn}, limits updates to input prompts~\cite{niu2024test}, or relies on inherently noisy zeroth-order gradient estimation~\cite{spall2002multivariate, deng2025test}, which limits their ability to resolve complex distribution shifts.
\ 2) high computational overhead: existing approaches~\cite{niu2024test, deng2025test} require numerous forward passes to match the performance of BP-based methods. Furthermore, these methods waste resources by performing adaptation on every batch indefinitely, regardless of domain stability.

To address these challenges in TTA, we introduce Projected Adaptation via Covariance Evolution (\methodname{}), an efficient BP-free adaptation system that expands learning capabilities while minimizing inference overhead.

We start by adapting the model utilizing the Covariance Matrix Adaptation Evolution Strategy (CMA-ES), since its effectiveness for TTA was already confirmed~\cite{niu2024test}. The high dimensionality of neural network weights makes direct CMA-ES optimization intractable. To circumvent this,~\cite{niu2024test} focuses on updating input prompts rather than internal weights. However, our experiments demonstrate that input prompt tuning is significantly inferior to updating the affine parameters of normalization layers (Fig.~\ref{fig:gt_finetune_lr}), as commonly done in TTA~\cite{niu2023towards,wang2021tent,eata}. While this finding motivates the application of CMA-ES directly to the normalization layers to maximize adaptation capability, doing so naively remains computationally intractable due to the high-dimensional nature of these parameters. Therefore, we exploit the observation of low intrinsic dimensionality for TTA gradients~\cite{duan2025lifelong}, a property we verify empirically in Fig.~\ref{fig:grad_pca}. \methodname{} optimizes a low-dimensional vector and projects it into the high-dimensional normalization weight space via the Fastfood transform~\cite{le2013fastfood}, enabling highly effective adaptation without backpropagation.

Furthermore, we observe that the CMA-ES adaptation predominantly occurs at the onset of a stable domain, yielding negligible accuracy improvements in later batches (Fig.~\ref{fig:stop_adapt}). We leverage this by designing a dynamic stopping criterion based on the mean shift of the CMA-ES distribution. Once adaptation halts, the model operates with almost zero computational overhead. When a domain shift is detected, adaptation automatically resumes. Finally, to accumulate knowledge across domain shifts, we maintain a bank of domain-specialized adaptation vectors, allowing the system to rapidly recover performance when encountering repeated domains. These components combine to form a fully practical TTA system, especially suitable for long-term deployment.

In summary, our contributions are:
\begin{itemize}
\item We develop \methodname{}, a BP-free continual TTA method that efficiently optimizes normalization layers using CMA-ES, utilizing Fastfood transform projections to search a low-dimensional subspace.

\item We introduce a dynamic stopping criterion based on the CMA-ES distribution mean that halts adaptation during stable domain phases, eliminating computational overhead.

\item We integrate a domain-specialized vector bank, enabling the model to accumulate knowledge and rapidly adapt to recurring domains.
\end{itemize}
\noindent These lead to our proposed method outperforming the best BP-free approaches in both efficiency and accuracy, achieving around 50\% reduction in runtime. Noticeably, PACE significantly surpasses the other CMA-ES-based method (FOA).

\section{Related Work}

\noindent \textbf{Test-Time Adaptation.} 
TTA targets distribution shifts during inference by adapting pre-trained models without access to source data or labels~\cite{wang2021tent}. 
Continual TTA further assumes a non-stationary environment where domains change without access to domain labels~\cite{wang2022continual}.
Existing methods can be divided into backpropagation-based (BP-based) and backpropagation-free (BP-free).

\smallskip
\noindent \textbf{BP-based methods.}
These methods utilize gradient descent to update model parameters at test time. 
To balance efficiency with learning capacity, most approaches restrict updates to affine parameters in normalization layers~\cite{niu2023towards, eata, wang2021tent, dobler2023robust}. 
In the absence of labels, these methods employ self-supervised objectives such as entropy minimization~\cite{wang2021tent,eata,tan2025uncertainty}, pseudo-labeling variants~\cite{dobler2023robust,wang2022continual}, rotation prediction~\cite{sun2020test}, or feature distribution alignment~\cite{mirza2022efficient}. 
Because self-supervised signals can be unreliable, stabilization techniques are often required. 
For instance, SAR~\cite{niu2023towards} filters samples by entropy and seeks flat loss minima, while CoTTA~\cite{wang2022continual} adopts a teacher-student framework with stochastic weight restoration. 
Despite their effectiveness, BP-based methods demand high memory and differentiable weights, which limits their utility on quantized or resource-constrained edge devices.

\smallskip
\noindent \textbf{BP-free methods.}
Early BP-free TTA focused on adjusting batch normalization statistics using test data batches~\cite{nado2020evaluating, schneider2020improving, gong2022note}. 
Subsequent work extended the statistics update to single-sample adaptation~\cite{khurana2021sita} and handling the temporal class correlation~\cite{gong2022note, zhao2023delta}. 
However, these approaches require the presence of batch normalization layers in model architectures.
Further works explored prototype-based classifier adjustment (T3A)~\cite{iwasawa2021test} and logits correction (LAME)~\cite{boudiaf2022parameter}.
None of those methods update core model weights, resulting in significantly limited learning capacity.
Recent work introduces weight-updating BP-free methods: FOA~\cite{niu2024test} adapts input prompts via CMA-ES, and ZOA~\cite{deng2025test} employs zeroth-order gradient estimation.
While promising, FOA's prompt-only updates restrict its flexibility. Furthermore, FOA adapts on every batch, demanding up to 28 forward passes, which is computationally prohibitive for every data sample in real-world deployment.
In contrast, ZOA reliably updates core model weights with only 2 forward passes. However, to remain competitive with BP-based methods, its computational demands are similar to FOA.
Our \methodname{} improves adaptation effectiveness through subspace adaptation of affine parameters of normalization layers via CMA-ES and minimizes overhead by introducing an automated stopping criterion that halts adaptation once it is no longer beneficial. 
Additionally, drawing on recent BP-based TTA work~\cite{vray2025reservoirtta}, we incorporate a domain-specialized vector bank to aggregate knowledge across diverse environments. This mechanism enables the model to rapidly recover performance when re-encountering domains, significantly enhancing its readiness for practical deployment.

\section{Problem Statement}
Continual TTA aims to adapt a pre-trained source model $f_{\bm{\theta}^{(0)}}$ to a non-stationary stream of unlabeled target data on-the-fly. The model is initially trained on a labeled source dataset $\mathcal{D}_s = \{(\bm{x}_i, y_i)\}_{i=1}^{N_s}$ sampled from distribution $\mathcal{P}_s$. At inference, the model sequentially encounters $A$ distinct target domains, $\mathcal{D}_t^{(1)}, \dots, \mathcal{D}_t^{(A)}$. For each domain $a \in \{1, \dots, A\}$, the unlabeled dataset is defined as:
\begin{equation}
\mathcal{D}_t^{(a)} = \{\bm{x}_j^{(a)}\}_{j=1}^{N_a}, \quad \text{where} \quad \bm{x}_j^{(a)} \sim \mathcal{Q}^{(a)}.
\end{equation}
The target domains are out-of-distribution ($\mathcal{Q}^{(a)} \neq \mathcal{P}_s$) and the sequence is non-stationary ($\mathcal{Q}^{(a)} \neq \mathcal{Q}^{(a+1)}$). The model must adapt online using only current samples, without access to $\mathcal{D}_s$ or past target data.
\section{Methodology}
\begin{figure}[!b]
    \centering
    \includegraphics[width=0.99\linewidth]{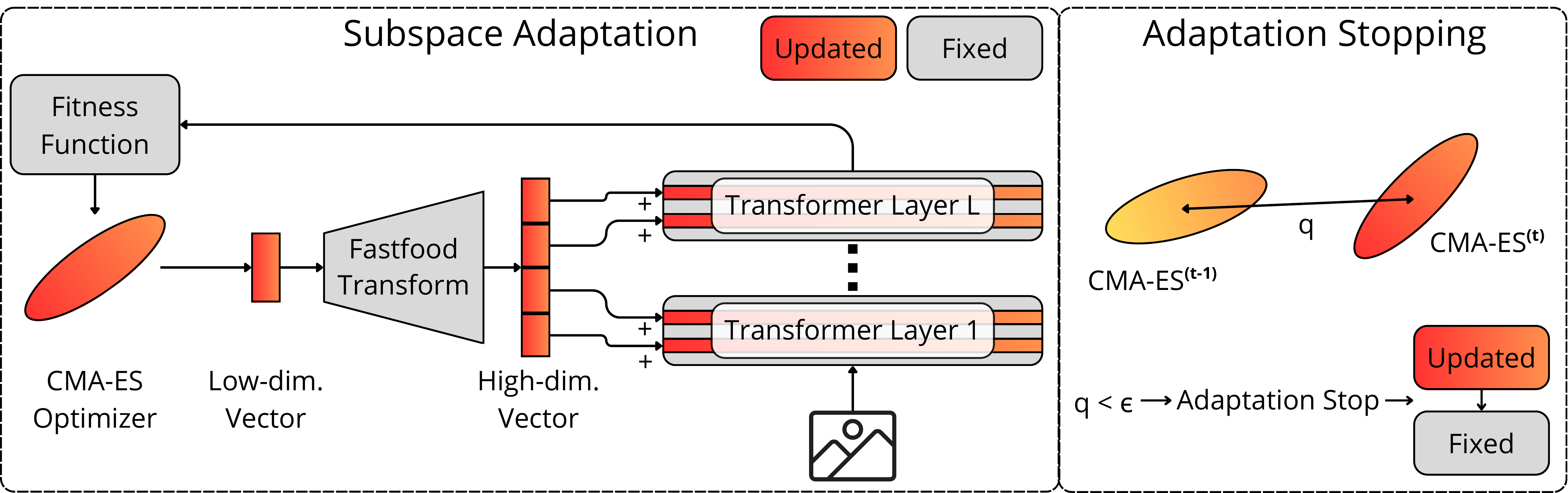}
    \caption{
    \textbf{Diagram of \methodname{}}.
    1)~\textbf{Subspace Adaptation}: we adapt the model by adding a high-dimensional random projection of a small, learnable vector to the model's normalization layer weights. We use the CMA-ES strategy to iteratively evolve a population of these vectors, selecting the one that minimizes the loss on current test samples.
    2)~\textbf{Adaptation Stopping}: for efficiency, we stop the adaptation when the mean of the distribution optimized by CMA-ES is lower than a threshold. Along with the \textbf{Domain-Specialized Vector Bank}, they make an effective and efficient TTA system.
    }
    \label{fig:diagram}
\end{figure}

We introduce a novel BP-free continual TTA method, named Projected Adaptation via Covariance Evolution (\methodname{}), designed to be a fully practical TTA system. It comprises three main components: Subspace Adaptation~\ref{sec:adaptation}, Adaptation Stopping~\ref{sec:stopping}, and Domain-Specialized Vector Bank~\ref{sec:buffer}. The diagram of our proposed \methodname{} method is presented at the end of this section (Fig.~\ref{fig:diagram}). 

\subsection{Subspace Adaptation}
\label{sec:adaptation}

\begin{figure}[!t]
    \centering
    \includegraphics[width=0.73\linewidth]{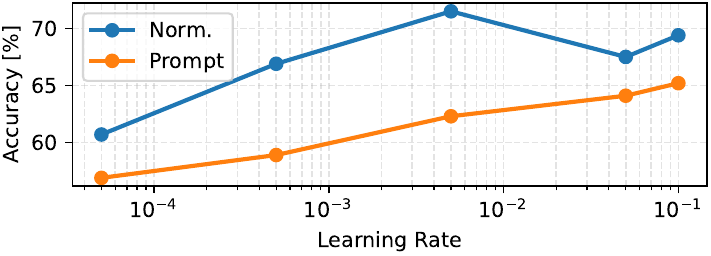}
    \caption{
    \textbf{Observed performance gap}. A comparison of updating the affine parameters of normalization layers (\textsc{Norm}.) versus three input prompts (\textsc{Prompt}) for a \mbox{ViT-B} model during test-time adaptation on \mbox{ImageNet-C}. We adapt these using ground-truth labels with SGD optimizer with varying learning rate. Updating the normalization layers allows the model to more effectively `correct' the covariate shift at each network depth for all reported learning rate values.
    }
    \label{fig:gt_finetune_lr}
\end{figure}

\begin{figure}[!t]
    \vspace{-1em}
    \centering
    \includegraphics[width=0.86\linewidth]{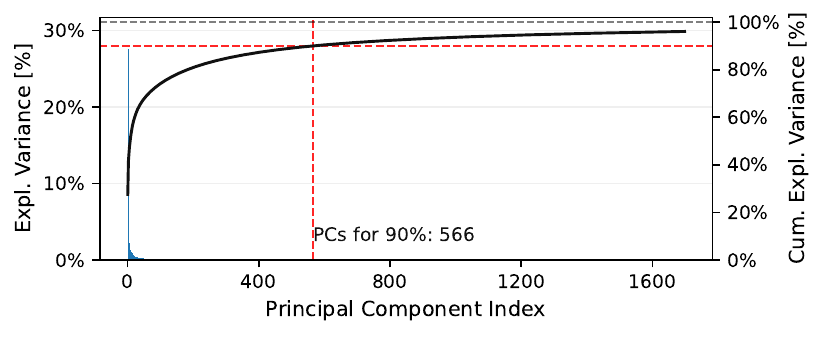}
    \caption{
    \textbf{Intrinsic dimensionality of continual TTA gradients}. The affine parameters of normalization layers from a \mbox{ViT-B} model are optimized via SGD with the loss function from FOA~\cite{niu2024test}. The concatenated gradients from all \mbox{ImageNet-C} domains reveal that only \num{566} components explain \SI{90}{\percent} of the variance. This highlights the low-dimensional nature of the adaptation space. Note that only the most significant components are shown for clarity. The analysis is based on \num{11729} gradient batches.
    }
    \label{fig:grad_pca}
\end{figure}

Building on established TTA frameworks~\cite{wang2021tent, niu2023towards, eata, dobler2023robust} and our observation that input prompt tuning is less effective than updating affine parameters of normalization layers (Fig.~\ref{fig:gt_finetune_lr}), we employ CMA-ES to optimize these layers. To ensure high-dimensional optimization remains tractable by CMA-ES, we leverage the observation that TTA gradients have low intrinsic dimensionality (as noted in~\cite{duan2025lifelong} and Fig.~\ref{fig:grad_pca}), which suggests that effective updates can be achieved within a low-dimensional subspace.
While this was also noticed by Duan~et~al.~\cite{duan2025lifelong}, using a few batches from a single domain, we extend this observation to all concatenated domains.
The primary directional components remain low-dimensional despite shifting domain distributions.
Although the optimization dynamics differ between SGD and CMA-ES, our analysis shows that the solution for the valid adaptation reside within the low-dimensional subspace. Consequently, we constrain CMA-ES to search for a solution exclusively within such a restricted subspace.

\smallskip
\noindent \textbf{Model update.}
We adapt the model weights by adding the constant random projection $\proj(\cdot)$ of a low-dimensional vector $\bm{v}\!\in\!\mathbb{R}^d$ into a high-dimensional space \mbox{$\proj(\bm{v})\in \mathbb{R}^D$}, where $D \gg d$. The adapted model weights are the affine parameters of normalization layers, therefore we set $D$ to be equal to their total dimensionality. We partition $\proj(\bm{v})$ to match the dimensionality of each parameter tensor and add the resulting components to the initial weights to get the adapted model $f_{\bm{\theta}^{(0)}+\proj(\bm{v})}$. Initializing with a zero vector ($\bm{v} = \bm{0}$) ensures the adaptation starts exactly from the state of the pre-trained model. Our adaptation objective is to find the optimal low-dimensional vector $\bm{v}^*$ that minimizes the fitness function $\mathcal{L}(\cdot)$ given test samples $\bm{x}$:
\begin{equation}
\bm{v}^* = \operatorname*{argmin}\limits_{\bm{v}} \mathcal{L}(f_{\bm{\theta}^{(0)} + \proj(\bm{v})}(\bm{x}))\,.
\end{equation}

\smallskip
\noindent \textbf{CMA-ES optimization.}
We propose to perform the adaptation using the CMA-ES strategy. Rather than directly optimizing the vector $\bm{v}$, CMA-ES maintains and optimizes a multivariate Gaussian distribution over the search space. At each iteration $t$ ($t$-th test batch), a population of $K$ candidate solutions is sampled from this distribution:
\begin{equation}
    \bm{v}_k^{(t)} \ \sim\  \bm{m}^{(t)} + \tau^{(t)} \mathcal{N}(\bm{0}, \bm{\Sigma}^{(t)})\,,
\end{equation}
where $k = 1, \cdots, K$. Here, $\bm{m}^{(t)}\!\in \mathbb{R}^{d}$ represents the mean of the search distribution at iteration step $t$, \,$\tau^{(t)}\!\in \mathbb{R}^{+}$ is the overall standard deviation controlling the global step size, and $\bm{\Sigma}^{(t)}$ is the covariance matrix defining the distribution's shape and orientation.

For each candidate $\bm{v}_k^{(t)}$, we evaluate its fitness by updating the model and computing the loss on the current test sample $\bm{x}^{(t)}$. This yields a fitness value $l_k$ for each candidate.
CMA-ES then updates the distribution parameters \mbox{($\bm{m}^{(t)}$, $\tau^{(t)}$, and $\bm{\Sigma}^{(t)}$)} for the next generation based on the ranking of the fitness values. This process systematically increases the likelihood of previously successful candidates (see~\cite{hansen2016cma} for details). Following \cite{niu2024test}, we output the prediction associated with the lowest fitness value as the final prediction of the model.

\smallskip
\noindent \textbf{Fitness function.}
We utilize the fitness function used in FOA~\cite{niu2024test} and ZOA~\cite{deng2025test} due to its proven effectiveness, and to enable direct comparison. Prior to adaptation, we pass a small set of source data through the model to calculate the means and standard deviations of activations from $L$ intermediate model blocks, denoted as $\{\mu_{s,i}, \sigma_{s,i}\}_{i=1}^L$. During test time, we calculate the corresponding statistics on the current batch of test samples $\bm{x}^{(t)}$, yielding $\{\mu_i(\bm{x}^{(t)}), \sigma_i(\bm{x}^{(t)})\}_{i=1}^L$. The fitness function combines the prediction entropy with the difference between these activation statistics:
\begin{equation}
\begin{aligned}
\mathcal{L}(f_{\bm{\theta}^{(0)}+\proj(\bm{v}_k^{(t)})}(\bm{x}^{(t)}))
\ = &\ \frac{1}{B \times C} \sum_{\bm{x} \in \bm{x}_t} \sum_{c=1}^C -y_c \log y_c \\
+ &\ \lambda \sum_{i=1}^{L} \left( \|\mu_i(\bm{x}^{(t)}) - \mu_{s,i}\|_2 + \|\sigma_i(\bm{x}^{(t)}) - \sigma_{s,i}\|_2 \right)
\end{aligned}
\end{equation}
where $B$ is the batch size, $C$ is the total number of classes, $y_c$ is the $c$-th element of the prediction probability vector \mbox{$\hat{\bm{y}} = f_{\bm{\theta}^{(0)}+\proj(\bm{v}_k^{(t)})}(\bm{x}^{(t)})$}, and $\lambda$ is a balancing hyperparameter.

\smallskip
\noindent \textbf{Efficient dimensionality expansion.}
To efficiently project low-dimensional vectors $\bm{v}$ into the high-dimensional parameter space, we employ the Fastfood transform~\cite{le2013fastfood}. Typically, dimensionality expansion requires multiplying $\bm{v}$ by a dense projection matrix $\mathbf{W} \!\in\! \mathbb{R}^{D \times d}$. Storing this dense matrix can incur high memory costs for large networks. Fastfood circumvents this bottleneck by approximating the dense Gaussian matrix $\mathbf{W}$ with a product of structured diagonal matrices and the Fast Walsh-Hadamard Transform (FWHT). We redefine the linear projection as:
\begin{equation}
    \mathbf{W}\bm{v} \approx \mathbf{S} \mathbf{H} \mathbf{G} \mathbf{P} \mathbf{H} \mathbf{B} \bm{v} = \proj(\bm{v})\,,
\end{equation}
where $\mathbf{B}\!\in\!\mathbb{R}^{d \times d}$ is a diagonal matrix with entries sampled uniformly from $\{-1, 1\}$, $\mathbf{P} \in \mathbb{R}^{D \times D}$ is a random permutation matrix, and $\mathbf{G} \in \mathbb{R}^{D \times D}$ is a diagonal matrix with entries drawn from a standard normal distribution $\mathcal{N}(0, 1)$. The matrix $\mathbf{S} \in \mathbb{R}^{D \times D}$ is a diagonal scaling matrix ensuring the rows of the resulting pseudo-random matrix possess the correct $L_2$ norm to approximate the $\chi$-distribution of a true Gaussian random matrix. Finally, $\mathbf{H}$ represents the Walsh-Hadamard matrix. We perform multiplication by $\mathbf{H}$ via the FWHT, entirely avoiding the instantiation of the matrix in memory.
We initialize the Fastfood transform components once at the beginning of the adaptation phase and freeze them for the remainder of the process.
The efficacy of our subspace adaptation relies on the premise that random linear projection from the low-dimensional search space $\mathbb{R}^d$ into the high-dimensional space $\mathbb{R}^D$ does not severely distort the optimization landscape. The Fastfood transform efficiently approximates a dense Gaussian random matrix~\cite{le2013fastfood}.
According to the Johnson-Lindenstrauss property of Gaussian random matrices~\cite{Johnson1984ExtensionsOL}, distances and geometric structures are preserved under this mapping with high probability.
This ensures that the geometric properties of the fitness landscape within the chosen subspace are sufficiently preserved, allowing the CMA-ES optimizer to navigate the reduced search space effectively.

By default, we set $d\!=\!2304$ and the dimensionality of the updated \mbox{ViT-B} model parameters is equal to $D\!=\!24576$. In that case, our Fastfood transform implementation significantly reduces memory overhead, requiring only \SI{0.5}{\mega\byte} compared to the \SI{216}{\mega\byte} needed for a dense projection matrix.

Implementation details regarding the Fastfood transform (Sec.~\ref{sec:add_details}) and insights regarding the subspace fitness landscape~(Sec.~\ref{sec:proj_type}) can be found in the Appendix.

\begin{figure}[!b]
    \centering
    \includegraphics[width=0.75\linewidth]{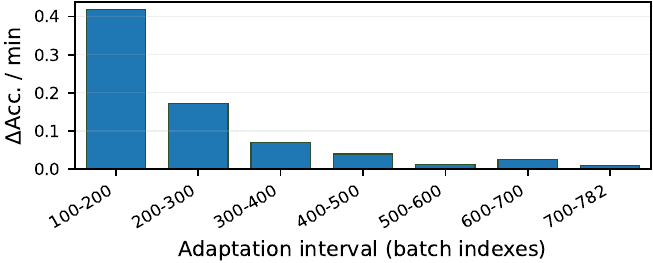}
    \caption{
    \textbf{Marginal accuracy gain per unit time} across adaptation budgets on \mbox{ImageNet-C} with \mbox{ViT-B} model. For each consecutive pair of adaptation step budgets, we compute the increase in mean accuracy across domains and divide it by the additional estimated wall-clock time required for that interval using our Subspace Adaptation with CMA-ES algorithm. The mean of CMA-ES distribution $\bm{m}$ is used as the evaluated model update. Higher bars indicate more efficient use of adaptation time, while decreasing bars indicate diminishing returns from further adaptation.
    }
    \label{fig:stop_adapt}
\end{figure}

\subsection{Adaptation Stopping}
\label{sec:stopping}
We observe that adaptation at the beginning of a stable domain drives the majority of performance improvements (Fig.~\ref{fig:stop_adapt}). However, the adaptation process incurs significant computational overhead. While test data distributions might remain stable for extended periods in real-world deployment, current BP-free TTA methods adapt indefinitely. This approach wastes computational resources and energy, a critical bottleneck for long-term deployments on embedded devices. To address this inefficiency, we introduce a heuristic to halt adaptation when it yields marginal performance gains. 

\smallskip
\noindent \textbf{Stopping heuristic.} CMA-ES continuously updates its distribution to track optimal candidates. Therefore, we monitor the relative change in distribution mean $\bm{m}$. Specifically, we stop the adaptation when the normalized difference between the distribution mean at the current time step $\bm{m}^{(t)}$ and the previous time step $\bm{m}^{(t-1)}$ falls below a convergence threshold $\epsilon$:
\begin{equation}
    \frac{\|\bm{m}^{(t)} - \bm{m}^{(t-1)}\|}{\|\bm{m}^{(t-1)}\|} < \epsilon\,.
    \label{eq:stopping_criterion}
\end{equation}

We rely on this specific formulation because the mean $\bm{m}^{(t)}$ represents the algorithm's current best estimate for the optimal parameters $\bm{v}$. When this relative change approaches zero, it indicates that the CMA-ES optimization has converged to a local optimum for the current data distribution. Furthermore, normalizing the difference by the previous mean ensures that our stopping criterion is scale-invariant, making it robust across different network layers or parameter magnitudes. 
Once the stopping criterion in Eq.~\ref{eq:stopping_criterion} is satisfied, the model is updated with the CMA-ES mean $\bm{m}^{(t)}$ and fixed.

\smallskip
\noindent \textbf{Resuming the adaptation.} 
Since continual TTA lacks explicit domain labels, we require a criterion to re-initialize the adaptation process. To detect these shifts, we adopt the domain shift detection scheme established in~\cite{hong2023mecta,chen2024cross,deng2025test}, which remains invariant to updates to the model.

We maintain the exponential moving average (EMA) of the activation statistics from the stem layer of the model:
\begin{equation}
\phi_{EMA}^{(t-1)} = \beta \phi^{(t-1)} + (1 - \beta) \phi_{EMA}^{(t-2)}\,,
\end{equation}
where $\phi^{(t-1)}$ represents the activation mean and variance for the $(t-1)$-th batch, and $\beta=0.8$ denotes the moving average factor. To quantify the shift $U$, we compute the symmetric Kullback-Leibler (KL) divergence between the current batch statistics $\phi^{(t)}$ and the historical EMA $\phi_{EMA}^{(t-1)}$:
\begin{equation}
U(\phi_{EMA}^{(t-1)}, \phi^{(t)}) = \frac{1}{H} \sum_{i=1}^{H} [ KL(\phi_{EMA}^{(t-1),i} \| \phi^{(t),i}) + KL(\phi^{(t),i} \| \phi_{EMA}^{(t-1),i}) ],
\end{equation}
where $H$ is the dimensionality of the statistics. A domain shift is detected when this distance exceeds a predefined threshold $\gamma$, at which point we re-initialize the adaptation using the Domain-Specialized Vector Bank. To ensure robust detection, we cease updating $\phi_{EMA}$ once the adaptation has stopped, by the technique described above.

\subsection{Domain-Specialized Vector Bank}
\label{sec:buffer}
In real-world, long-term deployments, domain recurrence can be a common phenomenon~\cite{vray2025reservoirtta}. A fully practical TTA system must account for this by rapidly reusing knowledge acquired from previously encountered domains.

To achieve this, we maintain a memory bank $\mathcal{B}$ of vectors $\bm{v}$ derived from past domains. Specifically, upon detecting a domain shift, we archive the current mean of the CMA-ES search distribution, $\bm{m}$, into the bank. For the incoming domain, the new CMA-ES optimizer is initialized using the archived mean that minimizes the fitness function $\mathcal{L}$ on the current data batch $\bm{x}^{(t)}$:
\begin{equation}
    \bm{m}_{init} = \underset{\bm{m}_i \in \mathcal{B}}{\arg\min} \; \mathcal{L}(f_{\bm{\theta}^{(0)} + \proj(\bm{m}_i)}(\bm{x}^{(t)}))\,.
\end{equation}
Adaptation then proceeds conventionally. This retrieval mechanism facilitates the rapid reuse of learned experiences, effectively preventing performance degradation even in the presence of sudden domain shifts.

To ensure strict memory bounds, we constrain the maximum capacity of the bank to $p$ vectors. When a newly optimized mean is added to a full bank~\mbox{($|\mathcal{B}| > p$)}, we employ a redundancy-based removal policy. We calculate the average pairwise cosine similarity for each mean $\bm{m}_i$ in the bank and discard the vector that exhibits the highest average similarity to the others:
\begin{equation}
    \bm{m}_{drop} = \underset{\bm{m}_i \in \mathcal{B}}{\arg\max} \; \frac{1}{|\mathcal{B}|-1} \sum_{\bm{m}_j \in \mathcal{B} \setminus \{\bm{m}_i\}} \frac{\bm{m}_i \cdot \bm{m}_j}{\|\bm{m}_i\| \|\bm{m}_j\|}
\end{equation}
This strategy effectively prunes the most redundant information, maximizing the diversity of the domain representations stored in the bank.

Subspace adaptation enables memory-efficient storage of knowledge from previous domains. A single \texttt{float32} domain vector with our default dimensionality ($d\!=\!2304$) occupies approximately \SI{0.0088}{\mega\byte} of memory. Consequently, the total memory usage is only about \SI{0.26}{\mega\byte} when the bank's maximum default capacity $p\!=\!30$ is utilized.

\section{Experiments}
\label{sec:experiments}
\noindent \textbf{Datasets and models.}
We conduct experiments on three standard TTA benchmarks: \mbox{ImageNet-C}~\cite{hendrycks2019robustness}, ImageNet-R~\cite{hendrycks2021many}, and DomainNet-126~\cite{peng2019moment}. \mbox{ImageNet-C} consists of 15 corruption functions across five severity levels. Following the protocol in~\cite{wang2022continual}, we evaluate using the classic corruption sequence with the highest severity level. ImageNet-R provides diverse renditions of \num{200} ImageNet classes, while DomainNet-126 contains images from four distinct domains (real, clipart, painting, and sketch).
Experiments are reported with both full-precision and quantized versions of \mbox{ViT-B}~\cite{dosovitskiy2020vit} and DeiT-B~\cite{pmlr-v139-touvron21a} models. Unless otherwise specified, experiments are reported with full-precision models. We implement quantization using PTQ4ViT~\cite{yuan2022ptq4vit}, following~\cite{niu2024test}. For ImageNet benchmarks, we use checkpoints trained on the ImageNet-1K~\cite{deng2009imagenet} training set obtained from the \texttt{timm} repository~\cite{rw2019timm}. For DomainNet-126, we utilize models trained on the \emph{real} domain using the repository from~\cite{zhang2023unified} and evaluate on the remaining three domains.

\smallskip
\noindent \textbf{Baselines.} 
We compare against both previously defined categories of state-of-the-art TTA methods: BP-free, including LAME~\cite{boudiaf2022parameter}, T3A~\cite{iwasawa2021test}, FOA~\cite{niu2024test}, and ZOA~\cite{deng2025test}, and BP-based, specifically TENT~\cite{wang2021tent}, CoTTA~\cite{wang2022continual}, and SAR~\cite{niu2023towards}. To ensure a fair comparison, we also compare with ZOA~($\text{fp}\!\!=\!\!28$), a modified version of ZOA where we match the number of forward passes used in FOA and \methodname{} by increasing the sampled perturbations for gradient estimation to \num{27}. Finally, NoAdapt represents the fixed source model without adaptation.

\smallskip
\noindent \textbf{Implementation details.} 
For all experiments, we maintain a batch size of \num{64} to ensure consistency with prior work~\cite{niu2024test,deng2025test,wang2022continual}. We adopt the hyperparameter settings specified in the original papers for all baselines. In the instances where they were not provided, the learning rate was specifically tuned for the corresponding experimental setup and model.
Following FOA, we set the optimization vector dimensionality $d$ to \num{2304} and configure the CMA-ES population size $K$ to \num{28} ($4 + 3 \times \log{(d)}$ as suggested in~\cite{hansen2016cma}).
While Fig.~\ref{fig:grad_pca} suggests a lower intrinsic dimensionality, we use $d=2304$ because our subspace relies on random projection rather than precise principal components. This over-parameterization increases the likelihood of intersecting with the effective optimization manifold.
Following both FOA and ZOA, we use the validation set of \mbox{ImageNet-1K} to compute the statistics of ID data and set the $\lambda$ to \num{0.2} for ImageNet-R and \num{0.4} for all other benchmarks. We set the domain shift detection threshold $\gamma$ to \num{0.03}, the adaptation stopping threshold $\epsilon$ is set to \num{0.045} and the maximum capacity of domain-specialized vector buffer $p$ to \num{30}. \methodname{} specifically updates the affine parameters of the normalization layers.

\smallskip
\noindent Further details and hyperparameter ablation studies are provided in the Appendix (Sec.~\ref{sec:add_details} and Sec.~\ref{sec:abl}, respectively).

\subsection{Results on Full-Precision Models}
\methodname{} achieves the highest average accuracy across all benchmarks and full-precision models, consistently outperforming other BP-free baselines (Tab.~\ref{tab:other_datasets}). Specifically, it surpasses direct competitors FOA and ZOA by \num{1.7} and \num{4.0} percentage points, respectively. When ZOA is modified to match the forward-pass budget of \methodname{} and FOA (ZOA ($fp\text{=}28$)), our method maintains a \num{1.4} percentage point lead despite adapting to fewer batches. 
While CoTTA achieves the highest accuracy on DomainNet-126 by leveraging backpropagation, it is significantly more memory-intensive than \methodname{} (Tab.~\ref{tab:performance_whole_inc}) and lacks support for quantized model updates. It is important to note that BP-based methods operate in a distinct category, as BP-free approaches are inherently constrained by the absence of direct gradient calculations. The fact that \methodname{} outperforms BP-based methods on average, despite these inherent limitations, further underscores the effectiveness of our approach.
More detailed results on \mbox{ImageNet-C} are presented in Sec.~\ref{sec:add_inc} in the Appendix.

\input{tables/other_datasets}
\input{tables/in-c_quantized}

\subsection{Results on Quantized Models}
We evaluate our approach against BP-free baselines for adapting quantized networks on \mbox{ImageNet-C} benchmark, with results summarized in Tab.~\ref{tab:quantized_results} for 8-bit and 6-bit \mbox{ViT-B} models. Our method consistently outperforms competing approaches across both bit widths. Notably, our 8-bit model achieves \SI{65.0}{\percent} accuracy, matching the 32-bit FOA baseline (\SI{65.0}{\percent}) while significantly exceeding almost all 32-bit competitors. 

The performance margin over ZOA ($\text{fp}\!=\!28$) increases as quantization becomes more aggressive, rising from \num{0.3} percentage points at 8-bit to \num{0.6} percentage points at 6-bit. Unlike ZOA, which requires bit-width-specific learning rates~\cite{deng2025test}, our approach maintains constant hyperparameters across all quantization levels, demonstrating superior robustness.

\subsection{Ablation Studies}
\noindent \textbf{Computational complexity analysis.}
Tab.~\ref{tab:performance_whole_inc} compares the wall-clock time, memory consumption and percentage of adaptation batches of \methodname{} against BP-based and BP-free baselines. While \methodname{} maintains the low memory footprint typical of BP-free methods, it improves efficiency over leading alternatives such as FOA and ZOA ($\text{fp}\!=\!28$). Specifically, our adaptation stopping technique reduces their wall-clock time by \SI{53}{\percent} and \SI{46}{\percent}, respectively, cutting runtime from over \SI{10}{\hour} to \SI{5.4}{\hour} while simultaneously increasing accuracy.

When we increase the threshold $\epsilon$ to match the runtime of default ZOA (\methodname{} with $\epsilon\!\in\!\{0.125, 0.14\}$), accuracy gains diminish to levels similar to ZOA or lower. However, \methodname{} achieves this performance by adapting to only \SI{3.5}{\percent}–\SI{4.8}{\percent} of batches, requiring only a single forward pass for the remaining samples. For \methodname{} ($K\text{=}6$), we reduce the population size and maintain the default $\epsilon$, which outperforms ZOA by \num{0.3} percentage points and reduces runtime by an additional \SI{0.1}{\hour}, despite adapting to \SI{13.7}{\percent} of batches. These results indicate that the adaptation stopping has its limits and keeping the $\epsilon$ threshold low while decreasing population size is a more effective strategy for drastically minimizing total runtime. Notably, ZOA cannot achieve similar efficiency gains because it requires at least two forward passes per batch for gradient estimation.

\input{tables/performance_whole_inc}

\smallskip
\noindent \textbf{Effectiveness of \methodname{} components.}
Tab.~\ref{tab:ablation} decomposes the performance gains of \methodname{} across its three core components. To isolate the impact of subspace adaptation, we first compare \methodname{}~v1 against FOA. While both methods employ CMA-ES, FOA updates input prompts whereas \methodname{}~v1 exclusively utilizes subspace adaptation. \methodname{}~v1 outperforms FOA by \num{1.1}~percentage points, demonstrating that the low-dimensional update of normalization layers is more effective than input prompt tuning.

Integrating the vector bank and shift detection (\methodname{}~v2) further improves accuracy. This suggests that the model effectively reuses fine-tuned vectors from previously encountered similar domains. This configuration achieves the highest overall result by maintaining updates for every data point. In contrast, \methodname{}~v3 prioritizes efficiency by utilizing adaptation stopping but not the vector bank, which reduces runtime. Ultimately, the full \methodname{} provides the optimal trade-off, combining all three components to achieve high accuracy with low computational~overhead.

\input{tables/ablation}

\begin{figure}[!ht]
    \centering
    \includegraphics[width=\linewidth]{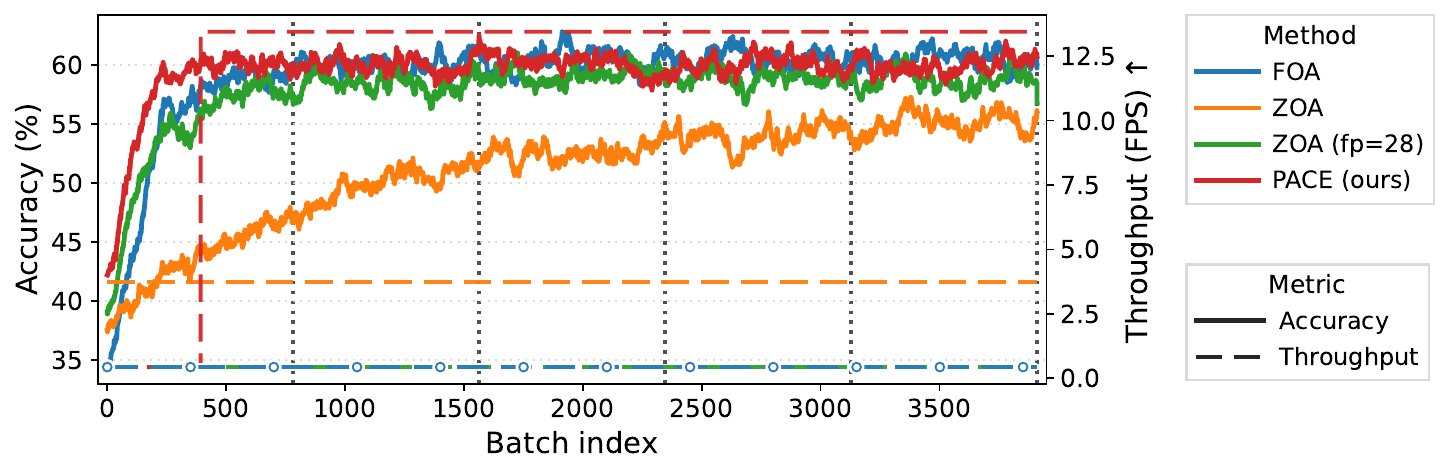}
    \caption{Smoothed per-batch accuracy for the repeated \emph{Glass Blur} domain from \mbox{ImageNet-C} with \mbox{ViT-B} model. The throughput was measured on a \textsc{Jetson Xavier NX}. The \textcolor{gray}{gray} vertical lines indicate the start of each repetition. The throughput from ZOA ($\text{fp}\!=\!28$) is covered by FOA.
    }
    \label{fig:repeated_glass_blur}
\end{figure}

\smallskip
\noindent \textbf{Performance on recurring domains.}
Tab.~\ref{tab:repeated_merge} compares \methodname{} against BP-free approaches in long-term continual adaptation scenarios involving recurring domains, following~\cite{niu2024test, vray2025reservoirtta, hoang2024persistent}. We evaluate performance on the repeated \mbox{ImageNet-C}. While competing methods require multiple passes over the benchmark to converge toward our accuracy, \methodname{} achieves a significantly high accuracy at the first round, then peaks by the second and maintains stability across the subsequent ones.
Only ZOA ($\text{fp}\!=\!28$) outperforms \methodname{} after five repetitions, gaining a marginal \num{0.2} percentage points of accuracy after the fourth repetition, while adapting to all test samples.
Fig.~\ref{fig:repeated_glass_blur} illustrates per-batch accuracy and throughput for repeated \emph{Glass Blur} domain from \mbox{ImageNet-C}. \methodname{} identifies when adaptation is no longer necessary and terminates redundant updates, maintaining a throughput of \SI{13}{\fps} on the \textsc{Jetson Xavier NX}. While FOA and ZOA ($\text{fp}\!=\!28$) maintain similar accuracy, they expend computational resources by adapting throughout the entire sequence, resulting in a significantly lower throughput of \SI{0.45}{\fps}.

\input{tables/repeated_merge}

\smallskip
\noindent \textbf{Subspace projection ablation.}
In Tab.~\ref{tab:projection}, we evaluate three distinct subspace projection methods within our framework: the Fastfood transform (\textsc{Fastfood}, our default configuration), a random dense projection matrix (\textsc{Dense}), and a data-driven matrix constructed from the dominant principal components of the SGD gradients gathered across the target domains (\textsc{PCA}), following the intrinsic dimensionality analysis from Fig.~\ref{fig:grad_pca}.
\input{tables/projection}

As expected, the performance of \textsc{Dense} is closely aligned with that of \textsc{Fastfood}, confirming that the Fastfood transform effectively approximates a dense random projection. Conversely, one might intuitively expect \textsc{PCA} to outperform these randomized approaches, given that its basis is constructed from the dominant gradient directions that minimize the loss function under standard SGD. However, this is not the case, and \textsc{PCA} results in inferior adaptation performance.
We hypothesize that this degradation stems from the \textsc{PCA} basis, which warps the fitness landscape into a severely ill-conditioned, anisotropic geometry. As demonstrated in~\cite{hansen2001completely}, CMA-ES suffers from delayed convergence on such ill-conditioned objective landscapes, as the algorithm requires a prolonged adaptation phase to align its mutation distribution with the underlying landscape contours. We explicitly confirm this severe ill-conditioning of the \textsc{PCA}-based landscape in the Appendix~(Sec.~\ref{sec:proj_type}).

\section{Conclusions}
In this work, we present \methodname{}, an efficient BP-free framework for continual TTA. By exploiting the low intrinsic dimensionality of normalization layer updates in TTA, \methodname{} overcomes the restricted learning capacity of prior BP-free methods without the prohibitive memory overhead of backpropagation. Furthermore, our adaptation stopping criterion and a domain-specialized vector bank ensure that \methodname{} remains efficient and robust during long-term deployment across recurring distribution shifts.
Consequently, this approach enables high-performance adaptation on resource-constrained edge devices.
Empirical evaluations show that \methodname{} establishes a new state-of-the-art for BP-free methods. These results also suggest that high-dimensional weight optimization is not a prerequisite for effective TTA. Instead, searching low-dimensional subspaces via evolution strategies provides a viable path toward practical, on-the-fly model refinement.

\section*{Acknowledgements}
This research was funded in whole or in part by National Science Centre, Poland, grant no \mbox{2024/53/N/ST6/03156} and \mbox{2023/51/D/ST6/02846}.  
We gratefully acknowledge Polish high-performance computing infrastructure PLGrid (HPC Center: ACK Cyfronet AGH) for providing computer facilities and support within computational grant no. \mbox{PLG/2025/018634} and \mbox{PLG/2025/018644}.
This research was funded in whole or in part by the Austrian Science Fund~(FWF) \mbox{10.55776/COE12}.
The authors would like to thank Prof. Piotr Skrzypczyński for his valuable feedback on this paper.

%
%
%
\bibliographystyle{splncs04}
\bibliography{bib}

\clearpage

\section*{Appendix}

\input{Supplementary/supplementary}
%
\end{document}

%% file: preamble.tex
\usepackage[dvipsnames,table]{xcolor}
\usepackage{multirow}
\usepackage{subcaption}
\usepackage{makecell}
\usepackage{paralist}
\usepackage{bm}
\usepackage{array}
\usepackage{graphicx}
\usepackage{booktabs}
\usepackage{amsmath}
\usepackage{amsfonts}
\usepackage{siunitx}
\usepackage{pifont}
\usepackage[nocompress]{cite}
\usepackage{hyperref}

\usepackage[ruled,vlined,linesnumbered]{algorithm2e}

\SetKwInput{Input}{Input}
\SetKwInput{Output}{Output}
\SetKwComment{Comment}{// }{ }

\newcommand{\methodname}{PACE}

\newcolumntype{C}[1]{>{\centering\arraybackslash}m{#1}}

\definecolor{greenx}{rgb}{0.0, 0.5, 0.0}

\definecolor{redx}{rgb}{0.84, 0.04, 0.33}

\DeclareMathOperator{\proj}{proj}
\DeclareSIUnit{\fps}{fps}

%% file: tables/other_datasets.tex
\begin{table}[t!]
    \caption{Accuracy on ImageNet-C (IN-C), ImageNet-R (IN-R) and \mbox{DomainNet-126} (DN-126) with ViT-B and DeiT-B. \textbf{Bold} indicates best result, \underline{underlined} second best.}
    \label{tab:other_datasets}
    \centering
    \resizebox{\linewidth}{!}{
  \begin{tabular}{lc@{\hskip 0.2in}>{\centering\arraybackslash}p{1cm}>{\centering\arraybackslash}p{1cm}>{\centering\arraybackslash}p{1.1cm}@{\hskip 0.3in}>{\centering\arraybackslash}p{1cm}>{\centering\arraybackslash}p{1cm}>{\centering\arraybackslash}p{1.1cm}@{\hskip 0.2in}>{\columncolor{black!8}}c}
 	 & & \multicolumn{3}{c}{ViT-B} & \multicolumn{3}{c}{DeiT-B} \\
 	 Method & Backprop. & IN-C & IN-R 
     & DN-126 & IN-C & IN-R & DN-126 & \textbf{Avg.} \\
    \midrule       
        NoAdapt & \ding{55} & 55.5 & 59.5 
        & 53.1 & 51.6 & 52.8 & 60.2 & 55.5 \\
        
        TENT & \ding{52} & 61.7 & \underline{63.9}  
        & 53.8 & 55.5 & 56.1 & 60.4 & 58.6 \\ 
        
        CoTTA & \ding{52} & 58.4 & 63.5  
        & \textbf{62.0} & 55.4 & 53.0 & 60.4 & 58.8 \\
        
        SAR  & \ding{52} & 61.5 & 63.3 
        & 53.8 & 59.4 & \underline{57.4} & 60.7 & 59.4 \\ 
        
        LAME & \ding{55} & 54.1 & 59.0 
        & 51.6 & 50.9 & 52.5 & 58.9 & 54.5 \\
        
        T3A & \ding{55} & 55.4 & 58.0 
        & 56.2 & 43.5 & 49.7 & 61.8 & 54.1 \\
        
        FOA & \ding{55} & 65.0 & 63.8 
        & 56.0 & 61.3 & 56.3 & 63.0 & 60.9 \\
        
        ZOA & \ding{55} & 61.5 & 60.7 
        & 55.8 & 56.9 & 53.5 & 62.4 & 58.5 \\

        ZOA ($\text{fp}\!\!=\!\!28$) & \ding{55} & \underline{66.3} & 62.6
        & 56.5 & \underline{61.7} & 55.8 & \underline{63.4} & \underline{61.1} \\

\midrule
\methodname{} (ours) & \ding{55} & \textbf{67.0} & \textbf{64.5} 
& \underline{57.0} & \textbf{62.7} & \textbf{59.5} & \textbf{64.3} & \textbf{62.5} \\  
	\end{tabular}
    }
\end{table}

%% file: tables/in-c_quantized.tex
\begin{table*}[t]
    \caption{Corruption Accuracy (\%) on ImageNet-C with \textbf{Quantized} ViT-B models. 
    }
    \label{tab:quantized_results}
    \centering
    \resizebox{\linewidth}{!}{
  \begin{tabular}{llccccccccccccccc>{\columncolor{black!8}}c}
 	\multicolumn{1}{c}{} & \multicolumn{1}{c}{}& \multicolumn{3}{c}{Noise} & \multicolumn{4}{c}{Blur} & \multicolumn{4}{c}{Weather} & \multicolumn{4}{c}{Digital} & \multicolumn{1}{c}{} \\
 	 Model & Method & Gauss. & Shot & Impul. & Defoc. & Glass & Motion & Zoom & Snow & Frost & Fog & Brit. & Contr. & Elas. & Pix. & JPEG & Avg. \\
    \midrule       
        \multirow{6}{*}{8-bit} & NoAdapt  & 55.8  & 55.8  & 56.5  & 46.7  & 34.7  & 52.1  & 42.5  & 60.8  & 61.4  & 66.7  & 76.9  & 24.6  & 44.7  & 65.8  & 66.7  & 54.1 \\  
        
        & T3A & 55.6 & 55.6 & 55.9 & 45.9 & 34.7 & 51.8 & 42.0 & 59.7 & 62.0 & 65.4 & 76.4 & 48.5 & 43.0 & 65.3 & 67.7 & 55.6 \\  
        
        & FOA & 60.7 & 61.4 & 61.3 & 57.2 & 51.5 & 59.4 & 51.3 & 68.0 & 67.3 & 72.4 & 80.3 & 63.2 & 57.0 & 72.0 & 69.8 & 63.5 \\  

        & ZOA & 57.9 & 60.6 & 61.4 & 51.2 & 44.6 & 56.6 & 51.8 & 64.6 & 61.8 & 63.6 & 78.1 & 54.5 & 54.4 & 68.7 & 68.3 & 59.7 \\  

        & ZOA ($\text{fp}\!\!=\!\!28$) & 60.4 & 62.6 & 63.3 & 56.8 & 53.6 & 62.1 & 58.5 & 67.2 & 67.0 & 70.4 & 79.4 & 60.8 & 63.7 & 72.2 & 72.2 & 64.7 \\ 

 & \methodname{} (ours) & 59.9 & 61.3 & 61.6 & 56.4 & 55.5 & 62.1 & 58.8 & 68.2 & 67.0 & 71.8 & 80.0 & 60.6 & 65.5 & 73.3 & 72.3 & 65.0 \\ 

\midrule       
\multirow{6}{*}{6-bit} &  NoAdapt  & 44.2  & 42.0  & 44.8  & 39.8  & 28.9  & 43.4  & 34.7  & 53.2  & 59.8  & 59.0  & 75.1  & 27.4  & 39.0  & 59.1  & 65.3  & 47.7 \\

& T3A & 42.9 & 40.6 & 42.1 & 31.4 & 25.5 & 40.2 & 31.9 & 48.7 & 58.3 & 58.1 & 73.8 & 27.2 & 36.7 & 58.8 & 65.7 & 43.3 \\ 

& FOA & 53.2 & 51.8 & 54.6 & 49.6 & 38.8 & 51.0 & 44.8 & 60.3 & 65.0 & 68.8 & 76.7 & 39.5 & 46.6 & 67.3 & 68.6 & 55.8 \\ 

& ZOA & 48.7 & 51.0 & 53.2 & 43.1 & 37.4 & 49.4 & 43.1 & 58.4 & 63.2 & 62.0 & 76.1 & 35.2 & 47.9 & 62.6 & 67.1 & 54.3 \\ 

& ZOA ($\text{fp}\!\!=\!\!28$) & 52.6 & 54.6 & 55.8 & 49.0 & 48.5 & 54.3 & 51.4 & 60.5 & 62.7 & 64.7 & 76.4 & 38.6 & 57.3 & 66.6 & 69.5 & 57.5 \\ 

 & \methodname{} (ours) & 53.2 & 53.9 & 55.0 & 49.2 & 48.6 & 55.3 & 50.8 & 61.0 & 64.2 & 66.4 & 77.0 & 42.2 & 56.8 & 68.1 & 69.6 & 58.1 \\
	\end{tabular}
	}
\end{table*}

%% file: tables/performance_whole_inc.tex
\begin{table}[t]
    \caption{Computation complexity comparison on \mbox{ImageNet-C} with \mbox{ViT-B}. Forward and backward passes (\,\#FP\,/\,\#BP\,) are counted for processing a single sample. The wall-clock time (hours) and memory usage (MB) are measured for processing the whole \mbox{ImageNet-C} on a single \textsc{RTX 4090 GPU}. Adapted batches are indicated by the percentage on which adaptation was performed.}
    \label{tab:performance_whole_inc}
    \centering
  \begin{tabular}{lccc@{\hskip 0.1in}>{\columncolor{black!8}}c@{\hskip 0.1in}ccc}
  Method & Backprop. & \#FP & \#BP & \makecell{Avg. Acc.\\(\%)} & \makecell{Runtime\\(hours)} & \makecell{Memory\\(MB)} & \makecell{Adapt.\\Batches (\%)}\\
  \midrule
      NoAdapt & \ding{55} &  1 & 0 & 55.5 & 0.01 & 819 & 0 \\
      TENT & \ding{52} &  1 & 1 & 61.7 & 0.03 & 5,165 & 100 \\
      SAR & \ding{52} & [1, 2] & [0, 2] & 61.5 & 1.1 & 5,166 & 100 \\
      CoTTA & \ding{52} & 3\,or\,35  & 1 & 58.4 & 1.5 & 16,836 & 100 \\
      T3A & \ding{55} & 1 & 0 & 56.9 & 0.7 & 957 & 100 \\
      FOA & \ding{55} & 28 & 0 & 65.0 & 11.6 & 832 & 100 \\
      ZOA & \ding{55} & 2 & 0 & 61.5 & 0.7 & 858 & 100 \\
      ZOA ($\text{fp}\!=\!28$) & \ding{55} & 28 & 0 & 66.3 & 10.0 & 862 & 100 \\
  \midrule
      \methodname{} ($\epsilon\!=\!0.125$) & \ding{55} & 1\,or\,28 & 0 & 61.5 & 0.8 & 863 & 4.8 \\
      \methodname{} ($\epsilon\!=\!0.14$) & \ding{55} & 1\,or\,28 & 0 & 61.0 & 0.7 & 863 & 3.5 \\
     \methodname{} ($K\!=\!6$) & \ding{55} & 1\,or\,6 & 0 & 61.8 & 0.6 & 863 & 13.7 \\
      \methodname & \ding{55} & 1\,or\,28 & 0 & \textbf{67.0} & 5.4 & 863 & 50.6 \\
	\end{tabular}
\end{table}

%% file: tables/ablation.tex
\begin{table}[!t]
    \caption{Ablation study on each component of \methodname{} on \mbox{ImageNet-C} with \mbox{ViT-B}.}
    \label{tab:ablation}
    \centering
  \begin{tabular}{l@{\hskip 0.1in}c@{\hskip 0.2in}c@{\hskip 0.2in}c@{\hskip 0.2in}>{\columncolor{black!8}}c@{\hskip 0.2in}c}
  Method & \makecell{adapt.\\stopping} & \makecell{vector\\bank} & \makecell{subspace\\adapt.} & \makecell{Avg.\\Acc.} & \makecell{Runtime\\(hours)}\\
  \midrule
      NoAdapt & & & & 55.5 & 0.01 \\
      Baseline (FOA) & & & & 65.0 & 11.6 \\
  \midrule
      \methodname{}~v1 & & & \ding{52} & 66.1 & 10.3 \\
      \methodname{}~v2 & & \ding{52} & \ding{52} & 67.2 & 10.3 \\
      \methodname{}~v3 & \ding{52} & & \ding{52} & 66.3 & 5.4 \\
      \methodname & \ding{52} & \ding{52} & \ding{52} & 67.0 & 5.4 \\
	\end{tabular}
\end{table}

%% file: tables/repeated_merge.tex
\begin{table}[b]
    \caption{Comparisons with state-of-the-art methods on \mbox{ImageNet-C} with \mbox{ViT-B} in long-term continual adaptation. We report average accuracy at each round of continual adaptation. \textbf{Bold} indicates best result, \underline{underlined} second.}
    \label{tab:repeated_merge}
    \centering
  \begin{tabular}{l@{\hskip 0.2in}c@{\hskip 0.2in}c@{\hskip 0.2in}c@{\hskip 0.2in}c@{\hskip 0.2in}c}
 	 Method & round 1 & round 2 & round 3 & round 4 & round 5 \\
    \midrule
        NoAdapt & 55.5 & 55.5 & 55.5 & 55.5 & 55.5 \\
        T3A & 55.4 & 55.9 & 55.2 & 55.0 & 54.6 \\
        FOA & 65.0 & 65.6 & 66.1 & 66.2 & 66.4 \\
        ZOA & 61.5 & 63.0 & 63.2 & 63.9 & 64.0 \\
        ZOA ($\text{fp}\!=\!28$) & \underline{66.3} & \underline{67.4} & \underline{67.3} & \underline{67.8} & \textbf{68.0} \\

\midrule
\methodname{} (ours) & \textbf{67.0} & \textbf{67.9} & \textbf{67.8} & \textbf{68.1} & \underline{67.8} \\ 

	\end{tabular}
\end{table}

%% file: tables/projection.tex
\begin{table}[!b]
    \caption{
    Accuracy (\%) of our method on ImageNet-C, evaluated across different subspace projection methods using a ViT-B model.
    }
    \label{tab:projection}
 \begin{center}
    \resizebox{0.6\linewidth}{!}{
  \begin{tabular}{l@{\hskip 0.1in}c@{\hskip 0.3in}c@{\hskip 0.3in}c@{\hskip 0.3in}c@{\hskip0.3in}}
   & \makecell{\textsc{Fastfood}\\(Def.)} & \textsc{Dense} & \textsc{PCA} \\
  \midrule
      \methodname{} (ours) & \textbf{67.0} & \textbf{67.0} & 64.8 \\
	\end{tabular}
	}
	 \end{center}
\end{table}

%% file: Supplementary/supplementary.tex
\renewcommand{\thefigure}{A.\arabic{figure}}
\setcounter{figure}{0}
\renewcommand{\thetable}{A.\arabic{table}}
\setcounter{table}{0}
\renewcommand{\thesection}{A.\arabic{section}}
\setcounter{section}{0}

\section{Overview}
This supplementary material provides comprehensive implementation details (Sec.~\ref{sec:add_details}) and additional experiments (Sec.~\ref{sec:add_exp}). Specifically, we present ablation studies on various hyperparameter values~(Sec.~\ref{sec:abl}), analysis of fitness landscape with different subspace projection approaches~(Sec.~\ref{sec:proj_type}), and detailed results on ImageNet-C benchmark~(Sec.~\ref{sec:add_inc}). 

\section{Implementation Details}
\label{sec:add_details}

We adopt the hyperparameter settings specified in the original papers for all baselines, except where they were not provided. In those instances, the learning rate was specifically tuned for our model and experimental setup. 
We utilized method implementations from the code repository of FOA~\cite{niu2024test} and ZOA~\cite{deng2025test}.  
In the following, we present the details regarding each method.

\noindent \textbf{\methodname{} (ours)}.
To ensure a fair comparison with FOA, we configure the CMA-ES population size $K$ to 28 and set the optimization vector dimensionality $d$ to 2304. Following FOA and ZOA, we utilize the validation set of ImageNet-1K to compute the statistics of in-distribution data, setting $\lambda$ to 0.2 for ImageNet-R and 0.4 for all other benchmarks. We set the domain shift detection threshold $\gamma$ to 0.03, the adaptation stopping threshold $\epsilon$ to 0.045, and the maximum capacity of the domain-specialized vector buffer $p$ to 30. Our method specifically updates the affine parameters of the normalization layers. Following ZOA, we keep the layer normalization parameters of the first block and the last three blocks of the tested models fixed.
Additionally, following FOA and ZOA, we utilize FOA's back-to-source activation shifting mechanism to shift the \texttt{CLS} features towards in-distribution domain.
In terms of Fastfood transform, to achieve the exact target dimensionality $D$, we pad the input to the nearest power of two ($C$) to enable the Fast Walsh-Hadamard Transform (FWHT), subsequently slicing the output to the required length. Moreover, because the utilized FWHT implementation~\cite{DaoFastHadamardSoftware} supports a maximum dimensionality of \num{32768}, configurations requiring $C > 32768$ cannot be processed natively in a single pass. 
When the target dimension exceeds this limit, we dynamically partition the operation into \num{32768}-sized sub-blocks. These are cross-mixed via a global permutation layer, using an empirical auto-calibration step at initialization to maintain orthonormality.

\noindent \textbf{FOA}~\cite{niu2024test}.
We set the number of input prompt embeddings to 3 and the population size $K$ to 28. In-distribution statistics are computed using the ImageNet-1K validation set. The loss trade-off parameter $\lambda$ is set to 0.2 for ImageNet-R and 0.4 for all other benchmarks, while the moving average factor for batch-to-source shift activation is maintained at 0.1.

\noindent \textbf{ZOA}~\cite{deng2025test}.
The learnable parameters are perturbed with a step size of 0.02 for gradient estimation, while the step size for the coefficients of different domain parameters is set to 0.05. The SGD optimizer with a weight decay of 0.4 is used to update the model parameters, and the AdamW optimizer with a weight decay of 0.1 is used to update the coefficients. 
The maximum number of domain knowledge parameters is set to 32. The learning rate for coefficients is set to 0.01 for all setups. The learning rate for model parameters is set to 0.0002 for 6-bit ViT-B and 0.0005 for all other models. In terms of ZOA ($fp=28$), the optimal learning rate was chosen on ImageNet-C and is set to 0.005 for all models and datasets.

\noindent \textbf{LAME}~\cite{boudiaf2022parameter}.
Following~\cite{niu2024test}, we use the kNN affinity matrix set to 5, as this value was found to be optimal for ImageNet-C.

\noindent \textbf{T3A}~\cite{iwasawa2021test}.
Following~\cite{niu2024test}, the number of
supports to restore $M$ is set to 20, as this value was found to be optimal for ImageNet-C.

\noindent \textbf{TENT}~\cite{wang2021tent}.
We use SGD optimizer, with a momentum of 0.9. The learning rate was tuned on ImageNet-C and set to 0.0001 for both ViT-B and DeiT-B models.

\noindent \textbf{SAR}~\cite{niu2023towards}.
We use SGD optimizer, with a momentum of 0.9. The learning rate was tuned on ImageNet-C and set to 0.001 for both ViT-B and DeiT-B models.
The entropy threshold $E_0$ is set to $0.4\times lnC$, where C is the number
of task classes.

\noindent \textbf{CoTTA}~\cite{wang2022continual}.
We use SGD optimizer, with a momentum of 0.9. The learning rate was tuned on ImageNet-C and set to 0.001 for ViT-B and 0.005 for DeiT-B.
The augmentation threshold $p_{th}$ is set to 0.1. The restoration probability is set to 0.01 and the EMA factor for teacher update is set to 0.999.

\section{Additional Experimental Results}
\label{sec:add_exp}
\subsection{Hyperparameter Values Ablation Studies}
\label{sec:abl}
\newcommand{\subtab}[2]{Tab.~\ref{#1}~(#2)}

\noindent \textbf{Optimization dimensionality ($d$).}\ \ 
\subtab{tab:params_ablation}{a} shows performance degradation when $d$ is either increased or decreased. The reduction in accuracy at lower dimensions indicates insufficient expressivity for the optimization updates. Meanwhile, the drop at higher dimensions is likely due to the increased search space complexity, which would require a larger CMA-ES population than our budget of \num{28}.
Although setting this parameter to 2304 yields state-of-the-art results on all tested benchmarks, it may require tuning for different neural network architectures.

\smallskip
\noindent \textbf{Domain-Specialized Vector Bank maximum capacity ($p$).}\ \ 
\subtab{tab:params_ablation}{b} shows the impact of the domain-specialized vector bank capacity $p$ on the ImageNet-C benchmark. Performance improves as $p$ increases, reaching a plateau at $p\!=\!15$. This behavior is expected since ImageNet-C consists of 15 distinct domains. As shown in Fig.~\ref{fig:inc}, our method successfully detects almost every domain transition and initializes from the bank (see Throughput). Consequently, increasing the capacity beyond the number of available domains yields no further gains.
The optimal value depends heavily on the test data.

\input{Supplementary/supp_tables/param_ablations}

\smallskip
\noindent \textbf{Domain shift threshold $\gamma$.}\ \ 
The effect of domain shift threshold~$\gamma$ is evaluated in \subtab{tab:params_ablation}{c}. While a lower $\gamma$ increases sensitivity to distribution shifts, frequent resets prevent the CMA-ES from reaching an optimal mutation distribution. On the other hand, a higher $\gamma$ results in insufficient sensitivity, leaving the model with suboptimal starting points when significant shifts occur. Our results suggest that a balanced threshold is necessary to maintain both detection accuracy and optimization stability.

\subsection{Fitness Landscape Analysis}
\label{sec:proj_type}
We analyze the low-dimensional fitness landscape across three distinct subspace projection methods within our framework: the Fastfood transform (\textsc{Fastfood}, our default configuration), a random dense projection matrix (\textsc{Dense}), and a data-driven matrix constructed from the dominant principal components of the SGD gradients gathered across the target domains (\textsc{PCA}), following the intrinsic dimensionality analysis from Fig.~3.
To do this, we employ a directional profiling approach similar to the linear path experiments in~\cite{Goodfellow2014QualitativelyCN}.
Specifically, we evaluate the local directional curvature of the fitness landscape $\mathcal{L}$ along the standard basis vectors $e_i \in \mathbb{R}^d$ of the low-dimensional subspace. Starting from the initial pre-trained source model weights $\theta^{(0)}$, we apply a continuous scalar perturbation $\alpha$ along a single coordinate axis. The corresponding high-dimensional weight configuration is parameterized as:
\begin{equation}
    \theta_i(\alpha) = \theta^{(0)} + proj(\alpha \cdot e_i)
\end{equation}
To approximate the local second derivative (the directional curvature $c_i$) at the origin ($\alpha = 0$), we utilize a central finite difference scheme with a fixed numerical step size $\delta > 0$ (empirically set to $\delta = 2$):
\begin{equation}
    c_i \approx \frac{\mathcal{L}(\theta_i(\delta)) - 2\mathcal{L}(\theta^{(0)}) + \mathcal{L}(\theta_i(-\delta))}{\delta^2}
\end{equation}
The overall landscape anisotropy is quantified by the empirical condition number~$\kappa$, defined as the ratio of maximum to minimum curvature across the subspace dimensions:
\begin{equation}
    \kappa = \frac{\max_{i} (|c_i|)}{\min_{i} (|c_i|)}
\end{equation}
\begin{figure}[!b]
    \centering
    \begin{subfigure}{1.\linewidth}
        \centering
        \includegraphics[width=\linewidth]{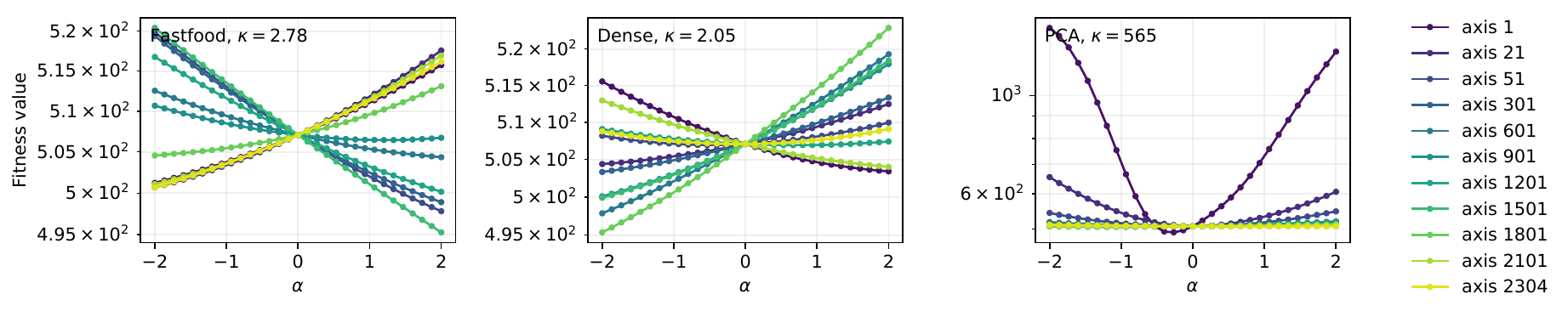}
        \caption{Gaussian Noise}
        \label{subfig:curv_gaussian}
    \end{subfigure}
    \begin{subfigure}{1.\linewidth}
        \centering
        \includegraphics[width=\linewidth]{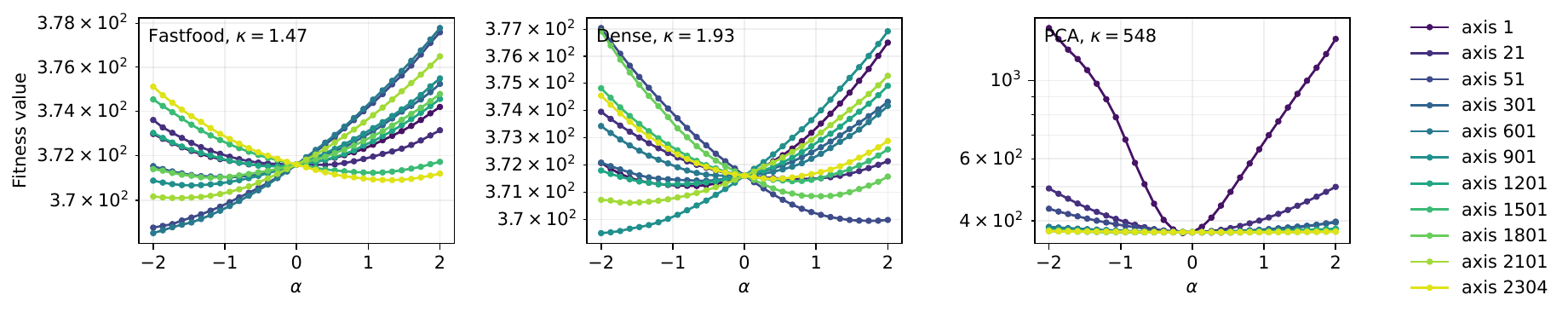}
        \caption{Snow}
        \label{subfig:curv_shot}
    \end{subfigure}
    \begin{subfigure}{1.\linewidth}
        \centering
        \includegraphics[width=\linewidth]{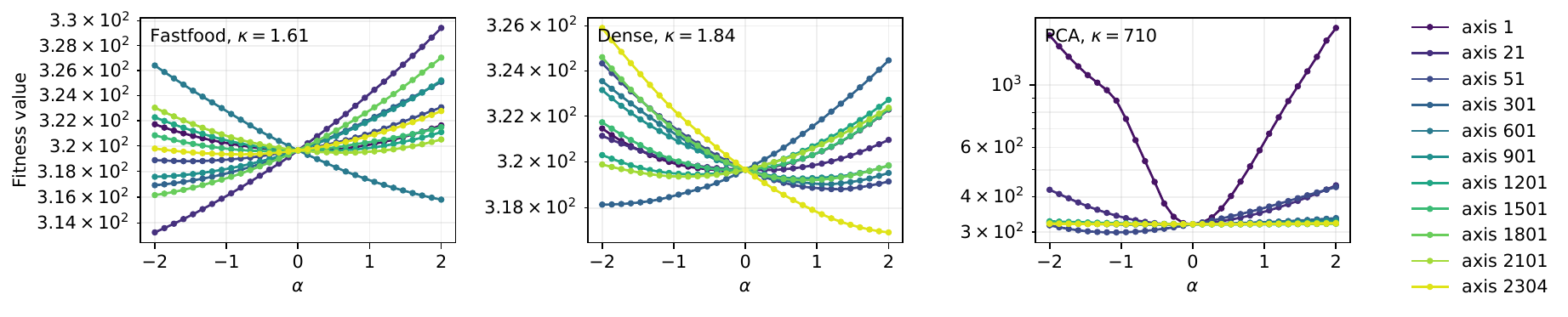}
        \caption{Pixelate}
        \label{subfig:curv_defocus}
    \end{subfigure}
    \caption{1D curvature profiles for \textsc{Fastfood} (\textbf{Left} panels), \textsc{Dense} (\textbf{Middle} panels), and \textsc{PCA} (\textbf{Right} panels) subspace projection methods evaluated on a ViT-B model across different ImageNet-C domains using the representative batches.}
    \label{fig:curvature}
\end{figure}
\begin{figure}[t]
    \centering
    \includegraphics[width=0.99\linewidth]{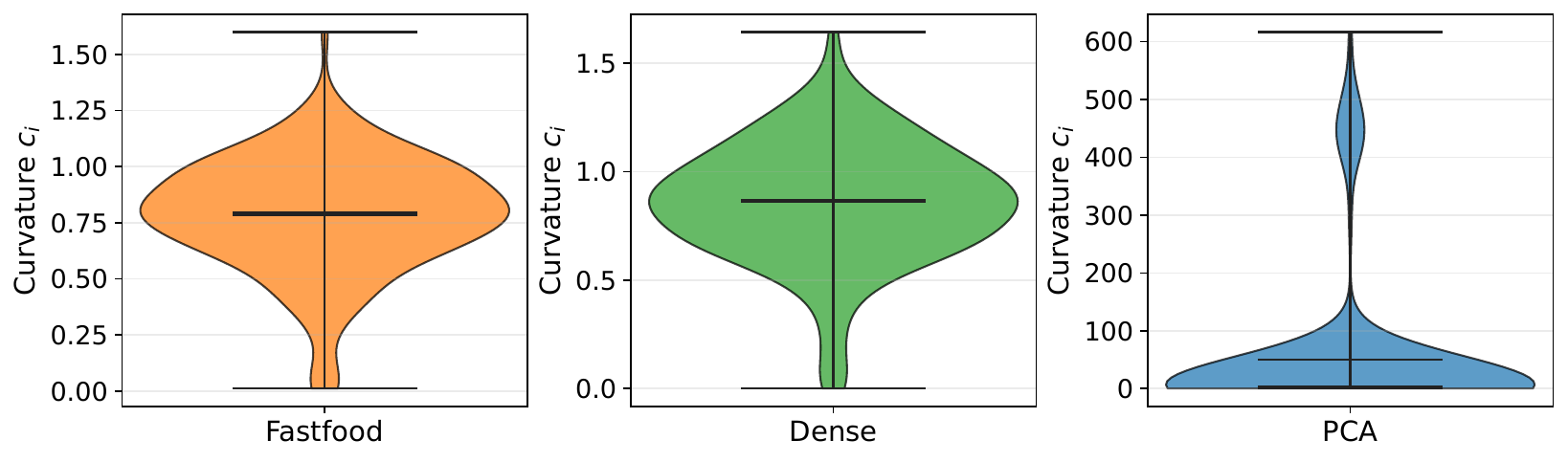}
    \caption{Distribution of 1D curvature estimates for \textsc{Fastfood}, \textsc{Dense}, and \textsc{PCA} subspace projection methods across all of the ImageNet-C domains and chosen axes on a ViT-B model. Horizontal markers indicate the mean and median.}
    \label{fig:curvature_violin}
\end{figure}

Fig.~\ref{fig:curvature} illustrates representative curvature profiles along chosen coordinate axes using a single batch from chosen domains of the ImageNet-C dataset, while Fig.~\ref{fig:curvature_violin} aggregates the global distribution of absolute curvatures across all evaluated target domains and subspace dimensions.
Aligning the transformation matrix with the dominant gradient eigenvectors (\textsc{PCA}) creates a highly ill-conditioned landscape relative to the randomized baselines (\textsc{Fastfood} and \textsc{Dense}). The empirical condition number of the \textsc{PCA} subspace is several orders of magnitude higher than the random projection configurations (e.g., $\kappa = 710$ for \textsc{PCA} versus $\kappa = 1.61$ for \textsc{Fastfood} on the Pixelate domain). Naturally, the primary \textsc{PCA} axes exhibit extreme curvature due to their alignment with the dominant gradient variance components, while the trailing dimensions rapidly flatten out. As demonstrated by the global curvature distributions in Fig.~\ref{fig:curvature_violin}, this long-tailed profile is consistent across all target domains under the \textsc{PCA} regime. In contrast, both \textsc{Dense} and \textsc{Fastfood} yield more uniform, symmetric curvatures that enable stable derivative-free optimization.

\subsection{Detailed ImageNet-C Results}
\label{sec:add_inc}
\begin{figure*}[!ht]
    \centering
    \includegraphics[width=1.0\linewidth]{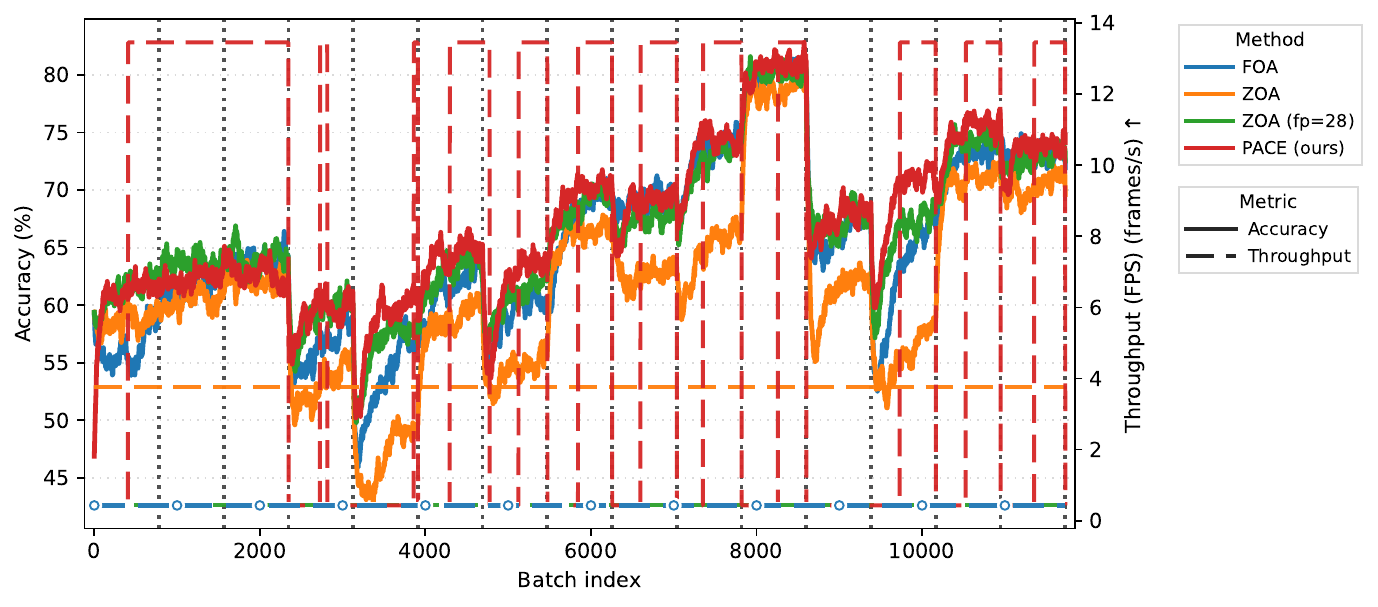}
    \caption{
    Smoothed per-batch accuracy for the ImageNet-C benchmark with ViT-B model. The throughput was measured on Jetson Xavier NX.
    The \textcolor{gray}{gray} horizontal lines indicate the start of the domain.}
    \label{fig:inc}
\end{figure*}

Fig.~\ref{fig:inc} illustrates the per-batch accuracy and throughput on the ImageNet-C dataset using the Jetson Xavier NX. As shown, \methodname{} dynamically modulates throughput by prioritizing adaptation at the onset of new domains (resulting in temporary throughput drops) and resuming high-speed inference once the domain stabilizes. In contrast, competing approaches maintain significantly lower throughput while achieving inferior accuracy (see Tab.~\ref{tab:imagenet-c-full-precision}). While certain domain shifts remain undetected due to inter-domain similarity, this suggests that re-initiating adaptation in such instances is unnecessary.

\input{Supplementary/supp_tables/in-c}

\clearpage

%
%
%
%

%% file: Supplementary/supp_tables/param_ablations.tex
\begin{table}[!t]
    \centering
    \caption{
    Accuracy (\%) of \methodname{} on \mbox{ImageNet-C} with a ViT-B model.
    }
    \label{tab:params_ablation}
    
    \vspace{1em}
    \textbf{(a) low-dimension space ($d$)}
    \smallskip
    
    \begin{tabular}{l@{\hskip 0.2in}c@{\hskip 0.2in}c@{\hskip 0.2in}c@{\hskip 0.2in}c@{\hskip 0.2in}}
     & 300 & 768 & 2304 & 3000 \\
    \midrule
      \methodname{} (ours) & 63.8 & 65.4 & \textbf{67.0} & 66.5 \\
	\end{tabular}

    \bigskip
    \textbf{(b) maximum vector bank capacity ($p$)}
    \smallskip
    
    \begin{tabular}{l@{\hskip 0.2in}c@{\hskip 0.2in}c@{\hskip 0.2in}c@{\hskip 0.2in}c@{\hskip 0.2in}c@{\hskip 0.2in}c@{\hskip 0.2in}}
     & 0 & 5 & 15 & \makecell{30} & 40 & 50 \\
    \midrule
    \methodname{} (ours) & 66.3 & 66.6 & \textbf{67.0} & \textbf{67.0} & \textbf{67.0} & \textbf{67.0} \\
	\end{tabular}

    \bigskip
    \textbf{(c) domain shift threshold ($\gamma$)}
    \smallskip

    \begin{tabular}{l@{\hskip 0.2in}c@{\hskip 0.2in}c@{\hskip 0.2in}c@{\hskip 0.2in}c@{\hskip 0.2in}c@{\hskip 0.2in}c@{\hskip 0.2in}}
     & 0.01 & 0.02 & \makecell{0.03} & 0.05 & 0.1 & 1.5 \\
  \midrule
      \methodname{} (ours) & 65.9 & \textbf{67.0} & \textbf{67.0} & 66.5 &  66.6 & 63.0 \\
	\end{tabular}

\end{table}

%% file: Supplementary/supp_tables/in-c.tex
\begin{table*}[b]
    \caption{Comparisons with SOTA methods on ImageNet-C with ViT-B regarding \textbf{Accuracy (\%)}.  \textbf{BP} is short for backward propagation.}
    \label{tab:imagenet-c-full-precision}
 \begin{center}
    \resizebox{1.0\linewidth}{!}{
  \begin{tabular}{lcccccccccccccccc>{\columncolor{black!8}}c}
 	\multicolumn{1}{c}{} & \multicolumn{1}{c}{}& \multicolumn{3}{c}{Noise} & \multicolumn{4}{c}{Blur} & \multicolumn{4}{c}{Weather} & \multicolumn{4}{c}{Digital} & \multicolumn{1}{c}{} \\
 	 Method & BP & Gauss. & Shot & Impul. & Defoc. & Glass & Motion & Zoom & Snow & Frost & Fog & Brit. & Contr. & Elas. & Pix. & JPEG & Avg. \\
    \midrule
NoAdapt & \ding{55} & 56.8  & 56.8  & 57.5  & 46.9  & 35.6  & 53.1  & 44.8  & 62.2  & 62.5  & 65.7  & 77.7  & 32.6  & 46.0  & 67.0  & 67.6  & 55.5 \\

TENT & \ding{52} & 57.6 & 59.8 & 60.9 & 51.2 & 49.4 & 59.6 & 53.2 & 64.0 & 62.7 & 67.8 & 78.6 & 66.5 & 54.5 & 70.0 & 69.7 & 61.7 \\   

CoTTA & \ding{52} & 57.4 & 58.4 & 59.7 & 47.5 & 38.3 & 54.9 & 47.3 & 62.4 & 63.4 & 69.9 & 77.8 & 54.3 & 47.8 & 68.0 & 68.6 & 58.4 \\

SAR  & \ding{52} & 59.1 & 61.1 & 61.6 & 54.2 & 55.1 & 58.6 & 55.7 & 60.3 & 61.5 & 64.3 & 76.6 & 58.2 & 58.1 & 68.6 & 68.9 & 61.5 \\ 

LAME & \ding{55} & 56.5 & 56.5 & 57.2 & 46.4 & 34.7 & 52.7 & 44.2 & 58.4 & 61.6 & 63.1 & 77.4 & 24.7 & 44.6 & 66.6 & 67.2 & 54.1 \\
T3A & \ding{55} & 56.4 & 56.6 & 56.7 & 45.5 & 34.4 & 51.9 & 43.4 & 60.6 & 62.8 & 60.9 & 77.1 & 45.8 & 44.5 & 66.7 & 68.5 & 55.4 \\ 

FOA & \ding{55}  & 56.3 & 61.7 & 63.9 & 56.1 & 53.3 & 61.4 & 58.9 & 67.5 & 69.1 & 73.0 & 80.2 & 65.9 & 62.1 & 72.5 & 73.0 & 65.0 \\

ZOA & \ding{55}  & 58.6 & 60.5 & 62.3 & 52.9 & 46.7 & 58.8 & 54.2 & 66.0 & 62.7 & 64.5 & 78.8 & 60.6 & 55.4 & 70.9 & 70.3 & 61.5 \\

ZOA ($fp\text{=}28$) & \ding{55} & 61.3 & 63.7 & 64.1 & 58.8 & 56.1 & 62.8 & 60.5 & 68.5 & 67.4 & 72.1 & 80.2 & 66.9 & 65.4 & 73.7 & 72.6 & 66.3 \\

\midrule
\methodname{} (ours)  & \ding{55} & 61.2 & 62.3 & 62.7 & 59.1 & 58.2 & 64.2 & 61.3 & 69.6 & 68.4 & 73.5 & 80.7 & 67.1 & 68.7 & 74.9 & 73.3 & 67.0 \\ 
	\end{tabular}
	}
  \end{center}
\end{table*}